\documentclass[10pt,twocolumn,letterpaper]{article}
\pdfoutput=1
\usepackage{cvpr}
\usepackage{times}

\usepackage{graphicx}

\usepackage[outdir=latex/images/]{epstopdf}

\DeclareGraphicsExtensions{.pdf,.png}

\usepackage{amsmath}
\usepackage{amssymb}

\makeatletter
\@namedef{ver@everyshi.sty}{}
\makeatother
\usepackage{tikz}
\usepackage{pgfplots}
\pgfplotsset{compat=1.17}

\usepackage[pagebackref=true,breaklinks=true,letterpaper=true,colorlinks,bookmarks=false]{hyperref}

\cvprfinalcopy 

\def\cvprPaperID{****} 

\begin{document}

\title{Semi-Supervising Learning, Transfer Learning, and Knowledge Distillation with SimCLR}

\author{Khoi Nguyen\\
Georgia Tech\\
Atlanta, Georgia\\
{\tt\small knguyen405@gatech.edu}
\and
Yen Nguyen\\
Georgia Tech\\
Atlanta, Georgia\\
{\tt\small yen.nguyen@gatech.edu}
\and
Bao Le\\
Georgia Tech\\
Atlanta, Georgia\\
{\tt\small bao.le@gatech.edu}
}

\maketitle

\begin{abstract}
Recent breakthroughs in the field of semi-supervised learning have achieved results that match state-of-the-art traditional supervised learning methods. Most successful semi-supervised learning approaches in computer vision focus on leveraging huge amount of unlabeled data, learning the general representation via data augmentation and transformation \cite{chen2020big}, creating pseudo labels \cite{sohn2020fixmatch}, implementing different loss functions, and eventually transferring this knowledge to more task-specific smaller models. In this paper, we aim to conduct our analyses on three different aspects of SimCLR, the current state-of-the-art semi-supervised learning framework for computer vision. First, we analyze properties of contrast learning on fine-tuning, as we understand that contrast learning is what makes this method so successful. Second, we research knowledge distillation through teacher-forcing paradigm. We observe that when the teacher and the student share the same base model, knowledge distillation will achieve better result. Finally, we study how transfer learning works and its relationship with the number of classes on different data sets. Our results indicate that transfer learning performs better when number of classes are smaller.

\end{abstract}

\section{Introduction, Background and Motivation}
\subsection{Background}

Often time in the real world, we do not have the luxury of having a fully labeled data set to work with. Many of such massive data sets have only a subset of them that are labeled. To combat this, research has been done in the sub-fields of self-supervised learning, semi-supervised learning and unsupervised learning with the goal of achieving model performance as effective as the traditional supervised learning methods, while leveraging huge amount of unlabeled data. Of these, SimCLR emerges to be the state-of-the-art semi-supervised learning framework that surprisingly matches other state-of-the-art supervised learning methods on images classification task. Motivated by its success, our team has chosen to study the SimCLR framework and its application via transfer learning and knowledge distillation methodologies.

\subsection{Motivation and Potential Impacts}

Labeling data is an expensive and time-consuming process. As a result, recent research in self-supervised learning has been focusing on teaching models the representational power of unlabeled data, without any human annotation. This can be done through a variety of methods, but the main theme revolves around learning through the augmentation and transformation of unlabeled data. SimCLR framework thrives on this concept by leveraging on training huge unlabeled dataset using contrastive learning. This method involves creating a pair of images that are transformed and augmented (cropping, flipping, color jittering and greyscaling) from an original image. Then, these pairs are fed into an encoder (e.g. a Resnet) to create image representation vectors. The special “NT-Xent” loss function \cite{chen2020big} will maximize the similarity between these two vectors (positive pairs), while also distinguish them from other non-related images (negative pairs). Consequently, original image features are extracted from these representation vectors, and are then fed to downstream linear classifiers for task-specific computation. This self-supervised pretraining process is task-agnostic. The goal here is to capture as much representational information as possible from the unlabeled dataset. The next step in the SimCLR framework involves fine-tuning the task-agnostic pretrained model into a task-specific one. This is to ensure that all the general representations that the model knows are fine-tuned to get familiar with the task at hand. Finally, the fine-tuned model can be used as a teacher to train a student network via transfer learning and knowledge distillation. This entire process not only helps us harvest useful information from huge unlabeled dataset where labeled data is limited, but also creates a pipeline to transfer those general knowledge to solve much more specific problems.


We believe that semi-supervised learning is incredibly applicable in many real-world applications as more unlabeled data being generated and labeling data getting pricier. For instance, in medical image diagnostic, it takes time for an experienced doctor to look through X-ray images and determine its label (making a diagnosis). In document classification, it is not feasible for a person to read through every line to determine which category that document belongs to. Fortunately, for any text related Deep Learning applications, BERT has come to the rescue. This framework is quite similar to SimCLR where the model is pretrained on a large corpus of text to obtain general knowledge, then fine-tuned to tackle downstream NLP specific tasks \cite{devlin2018bert}. Additionally, when combined with transfer learning and knowledge distillation, semi-supervised learning opens the door to many mobile Deep Learning applications where having a big model is not feasible. A few examples of these include intelligent virtual assistants such as Cortana, Siri, Alexa, and Google Assistant \cite{Henderson2020ConveRTEA}.  


\subsection{Datasets}
Our experiments were conducted using mainly models that are pre-trained and fine-tuned on the ImageNet dataset \cite{sky2015imagenet}. This dataset consists of 1000 classes, with a total of 14,197,122 annotated images. We trust that this dataset serves us well in providing a generalized pre-trained models for us to run experiments on. 

Besides ImageNet, we also used Tensorflow datasets, including Cats and dogs \cite{parkhi2012catsanddogs}, Oxford 102 Flowers \cite{nilsback2008flowers}, the Describable Textures Dataset (DTD) \cite{cimpoi2014dtd},  CIFAR-100 \cite{Krizhevsky2012cifar100} and Food-101 dataset \cite{bossard2014food101}.

\section{Approach}


To understand semi-supervised learning thoroughly, we review recent proposed methods. We then focus on SimCLR: a framework based on contrast learning. We conjecture about the advantages of SimCLR specifically and contrast learning in general.

We examine common transfer learning and knowledge distillation methods. Big model produces state-of the art results, but it is not practical due to the model's big size. Transfer learning and knowledge distillation can solve this problem. Knowledge are transferred and distilled to create more compact models taking advantages of bigger or teacher models. Approaches such as tuning and teacher forcing are proven to produce good results.

We conjecture about:
\begin{itemize}
  \item impact of data pre-processing, augmentation.
  \item approaches to design student models to best utilize teacher forcing.
  \item effect of a dataset's number of classes on transfer learning results.
  \item effect of data transformation on transfer learning.
\end{itemize}

We anticipate the following problems:
\begin{itemize}
    \item Training SimCLR requires significant computing resources.
    \item Student model may not converge as expected.
\end{itemize}

\section{Contrast Learning and SimCLR}

\subsection{Contrast Learning}

There are two general approaches to learn visual representations:

\begin{itemize}
    \item learn to re-create an existing visual representation
    \item learn to discriminate, compare, or contrast two or more visual representation
\end{itemize}

We conjecture that the latter is easier. Intuitively, it may be easier to tell if two images are different. On the contrary, it seems difficult to describe an image in details and re-create it \cite{chen2020simple}. Also, image re-creation, which is computationally expensive, is not necessary for a compare-contrast task.

\subsection{SimCLR}

SimCLR follows the contrast learning path. With contrast learning, the task is simple: given a pair of images, can you simply answer whether they are similar or not?

To help answer the question above, a SimCLR model needs to be taught about similarity. Two versions of an image are created by some transformations. And the model is trained to recognize that the two images above are similar. This is positive pair. On the other hand, the model also should be able to tell if two images are not similar (they are created from different images). This is negative pair.

In essence, SimCLR performs the following:

\begin{enumerate}
  \item Apply data augmentation on every training images. This results in 2 versions (X1 and X2) for each original image X.
  \item Extract visual representations of the transformed images via encoders.
  \item Identify X2, given X1. 
\end{enumerate}

Step 3 requires a contrast loss function, which is defined in \cite{chen2020simple}.

\subsubsection{Data Augmentation}

Data augmentation is important for visual generalization and SimCLR's performance. It is shown that a careful selection of augmentation operations plays critical role in the recognition task outcome \cite{chen2020simple}. With different combination of augmentation operations, \cite{chen2020simple} shows that an order combination of random cropping and random color distortion yields the best result. Without color distortion, the model tends to over-fit due to color familiarity.

\subsubsection{Semi-Supervised Learning and Model Architecture}

Semi-supervised learning gains better performance when the model is big \cite{chen2020big}. As the model becomes bigger, it extracts better visual representation from unlabeled data. \cite{chen2020big} shows that this is especially true when the number of unlabeled data is dominant in the data set.

\section{Transfer Learning and Knowledge Distillation}
\subsection{Transfer Learning}

Transfer learning is the idea of utilizing knowledge acquired from previously learned task or domain using pre-trained models to solve new tasks and domains. The key motivation for transfer learning is the prohibitive costs of training large models which requires large data and computing resources. It might be impossible to get such large training data for every domain. 

In transfer-learning, a model is first trained on a large dataset such as ImageNet \cite{sky2015imagenet}. These pre-trained weights are then used to initialize a model where most of the lower layers are frozen, and one or more new layers are added on top for learning from scratch or fine-tuned using the target dataset. The reason is that most of the lower layers of a neural network would contain the learned representation of the large general dataset, whereas only the top layers are highly specialized for the task at hand. This set-up helps the model converge faster while requiring less training time and still achieve high accuracy. Transfer learning has proven to be highly successful in many vision-based tasks including object recognition \cite{Razavian2014CVPR}, object detection and segmentation as well as emerging domains such as medical imaging \cite{arora2021covid}. 

SimCLR outperforms the state-of-the-art when used as part of a transfer learning framework using fine-tuning for semi-supervised image classification. 

\subsection{Knowledge Distillation}

While transfer learning focuses on transferring the weights of a big network to another big network, knowledge distillation aims to transfer the representational learning from a large and deep neural network (the teacher model) to a smaller network (the student model). If knowledge distillation is widely successful, neural networks could potentially be applied in many industry applications.

The key idea of knowledge distillation is to use soft probabilities of the teacher network (\'teacher's logits\')  to supervise the training of the student network in addition to available class labels \cite{ba2014deepnets}. Intuitively, we may think that these soft probabilities reveal more information that the teacher has discovered than just the class labels alone. Similarly, a student produces a softened class probability distribution. The Loss function then is the linear combination of the knowledge distillation loss (compared to teacher's logits), and the usual cross entropy loss (compared to ground truth label) \cite{hilton2015distilNN}.

In SimCLR as well as in our project, however, the student network is trained on unlabelled data. So it is relying entirely on teacher's logits. The loss function is just the knowledge distillation loss. 

\section{Implementation}

\subsection{How We Utilized Existing Implementations}

We utilized the published pre-trained and fine-tuned SimCLR model to train different datasets for our transfer learning experiments.

\subsection{How We Improved Existing Implementations}

We had to re-implement and improved existing fine-tuning and knowledge distillation implementations to work for our experiments:

\begin{itemize}
  \item We updated fine-tuning implementation to process data with augmentation operations before training.
  \item We updated knowledge distillation implementation with different student architectures, such as ResNet50 and VGG16, fine-tuned teacher, and performed hyper-parameter fine-tuning.
\end{itemize}

\subsection{Code Repositories}

\begin{itemize}
  \item \href{https://github.com/google-research/simclr}{SimCLR official implementation}
  \item \href{https://github.com/knguyen100000010/semi-supervised-learning}{Our project GitHub}
\end{itemize}

\section{Experiments and Results}



\subsection{\textbf Data Augmentation Effect on Fine-tuning}

In this experiment, we want to evaluate the effect of data augmentation operations on SimCLR model during fine-tuning.

\subsubsection{\textbf Experiment Setup}
\begin{figure}[h]
\centering
\includegraphics[scale=0.23]{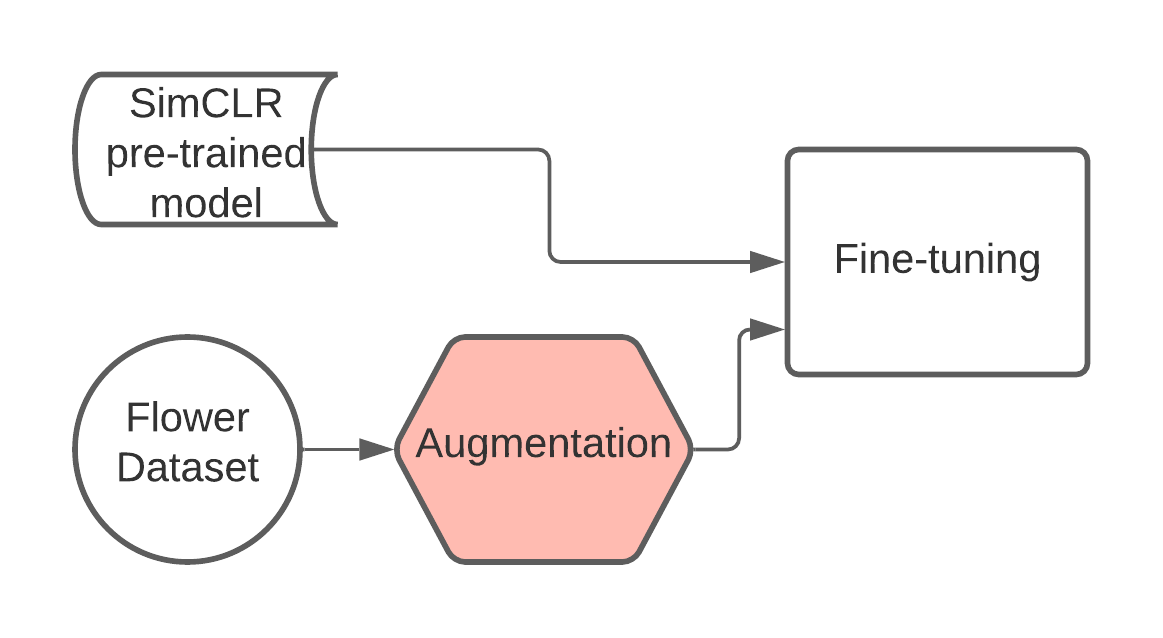}
\caption{Data augmentation operations are applied to Flower data set before being used for fine-tuning SimCLR pre-trained model}
\label{experiment1-setup}
\end{figure}

The experiment is setup like following, as depicted in figure \ref{experiment1-setup}:

\begin{enumerate}
  \item Load SimCLR model which was pre-trained on ImageNet dataset and fine-tuned with 1\% of labeled data. It should be noted that this pre-trained model is not familiar with the data set we are going to use for fine-tuning - which is Tensorflow Flower data set.
  \item We train the model with Tensorflow Flower data set to make the model familiar with flowers.
  \item We apply different combinations of data augmentation operations on the training data set and observe the outcome.
  \begin{itemize}
    \item Cropping only
    \item Flipping only
    \item Color distortion only
    \item Flipping \& Cropping
    \item Cropping \& Color distortion
    \item No augmentation
  \end{itemize}
  \item We hypothesize about the effect of data augmentation operations on SimCLR models during fine-tuning.
\end{enumerate}

The augmentation operations used for training the SimCLR model are:
\begin{itemize}
    \item Random cropping
    \item Random flipping
    \item Color distortion
\end{itemize}

SimCLR takes contrast learning approach to learn visual representations from unlabeled data. SimCLR pre-trained models are trained with unlabeled data which are heavily augmented with different set of operations. These operations include spatial ones, such as flipping and cropping, and appearance ones, such as color distortion \cite{chen2020simple}.

Augmentation operations are especially important for SimCLR algorithm. In fact, augmentation is the heart of this approach. For every image, two copies are created via augmentation operations. These co-related augmented images are critical for SimCLR model to learn the similarity and difference between them. Without heavy augmentations, the model cannot learn good representation of unlabeled images, so generalization will be poor when applied to unseen data set.

\subsubsection{\textbf Hypothesis}

We conjecture that data augmentation is still critical to fine-tuning. In other words, in this experiment, we show that SimCLR pre-trained model performs better when the training data set (Flower in this case) are augmented with similar data augmentation operations that were used for training the pre-trained model \cite{chen2020simple}. It should be noted that the SimCLR pre-trained model used in this experiment was trained with \textbf{unlabeled} data in a \textbf{task-agnostic} way. And the unlabeled data used for training the model go through \textbf{similar} augmentation operations \cite{chen2020simple}.

\subsubsection{\textbf Experiment Results and Analysis}

Our experiment results support this conjecture. For example, we observe that:
\begin{itemize}
    \item with individual operations, such as cropping, flipping, or color distortion, the accuracy is about the same when each operation is applied separately.  
    \item with a combination of two or more operations, the accuracy improves.
    \item the accuracy is highest when all operations mentioned above are applied.
    \item the accuracy is worst when no operation is used.
\end{itemize}
This trend is recorded in table 1 and figure 2.

\cite{chen2020simple} shows that with this combination, in order, the model learns the best representation of the unlabeled data. In this experiment, we see that it is also applicable for the fine-tuning process.

\begin{table}[h!]
\centering
\begin{tabular}{|c | c|} 
 \hline
 \textbf{Operations} & \textbf{Accuracy} \\ [0.5ex] 
 \hline\hline
 Cropping only & 0.84                    \\
 Flipping only & 0.841                   \\
 Color distortion only & 0.839           \\
 Cropping + Color distortion & 0.87      \\
 Flipping + Cropping & 0.873             \\
 All & 0.898                             \\
 None & 0.784                            \\ [1ex]
 \hline
\end{tabular}
\caption{Effect of data augmentation operations on SimCLR fine-tuning}
\label{table:1}
\end{table}

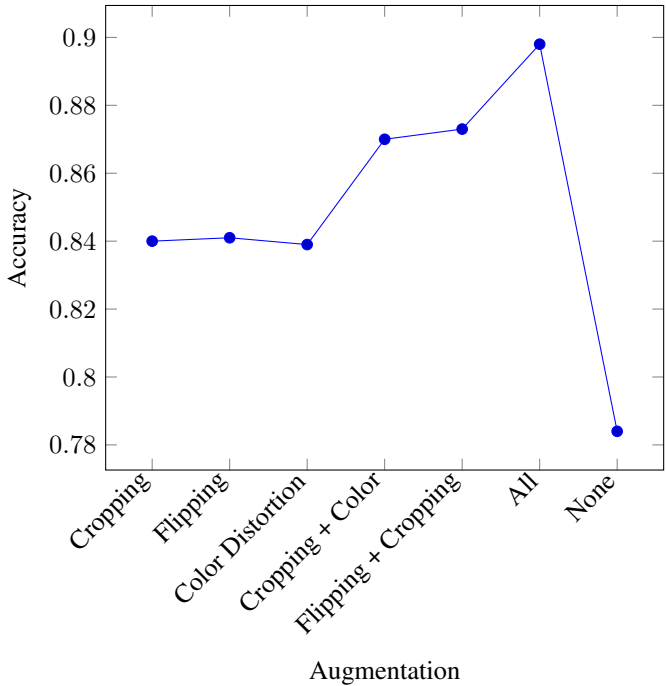
\begin{figure}[h]
\centering
\begin{tikzpicture}
    \begin{axis}
        [
        ,width=9cm
        ,xlabel=Augmentation
        ,ylabel=Accuracy
        ,xtick=data,
        ,xticklabels={
            Cropping,
            Flipping,
            Color Distortion,
            Cropping +  Color,
            Flipping + Cropping,
            All,
            None}
        ,x tick label style={rotate=45,anchor=east}
        ]
        \addplot+[sharp plot] coordinates
            {(0,0.84)
            (1,0.841)
            (2,0.839)
            (3,0.87)
            (4,0.873)
            (5,0.898)
            (6,0.784)
            };
    \end{axis}
\end{tikzpicture}
\caption{Comparison of data augmentation operations on SimCLR fine-tuning}
\end{figure}

\subsection{Knowledge Distillation with SimCLR via Teacher-Forcing Paradigm}

\begin{figure}[h]
\centering
\includegraphics[scale=0.4]{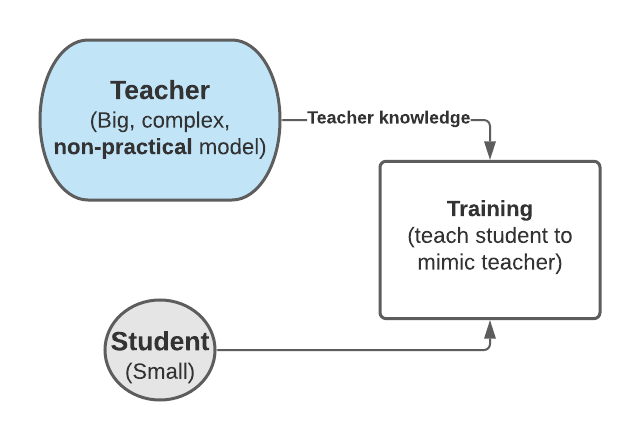}
\caption{Teacher-student: teacher's knowledge is distilled and transferred to student.}
\label{fig:augmentation}
\end{figure}

Models described in the literature achieve state-of-the-art accuracy. However, they are not practical due to their large size. For example, MoCo has about \textit{400} millions parameters \cite{chen2020simple}. These big models do not easily fit in user devices, such as mobile phones.

Knowledge distillation is an approach that can mitigate this issue. Knowledge distillation uses teacher-forcing method to distil the knowledge of larger model and transfer that knowledge to a much smaller, compact model (depicted in figure \ref{fig:augmentation}). The larger model plays teacher role, and the smaller one plays student role \cite{lucas2021distillation}.

With teacher-forcing, student's prediction is forced or encouraged to match teacher's prediction via a loss function. This is especially good for use-case where the number of unlabeled data is dominant. If the number of \textbf{labeled} data is quite large, we can create a loss function which takes into account teacher's prediction and ground truth labels \cite{chen2020big}. 

\begin{figure}[h]
\centering
\includegraphics[scale=0.375]{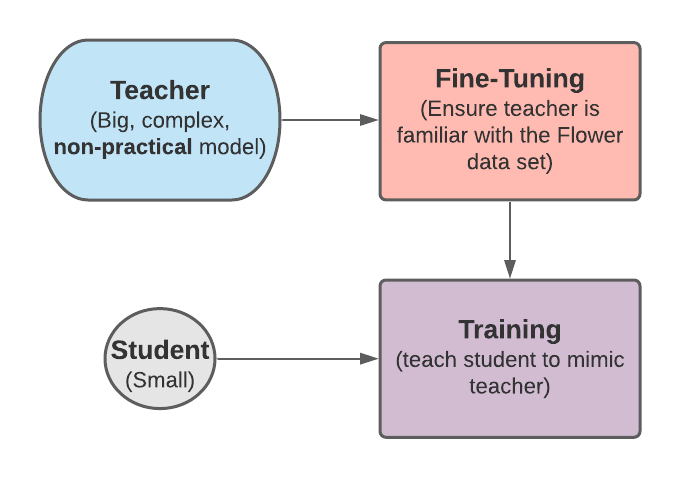}
\caption{Knowledge distillation - teacher forcing experiment setup: teacher (pre-trained SimCLR model) is fine-tuned with Flower dataset before being used for training student model.}
\label{fig:knowledge-distillation}
\end{figure}

\subsubsection{\textbf Experiment Setup}

The experiment setup is depicted in figure \ref{fig:knowledge-distillation}:
\begin{enumerate}
  \item The teacher model (pre-trained SimCLR model) is fine-tuned with Flower dataset to ensure it's familiar with the task.
  \item The student model is trained to mimic the teacher's predictions.
  \item We change the student model's architecture and observe the effect.
\end{enumerate}

\begin{figure}[h]
\centering
\includegraphics[scale=0.375]{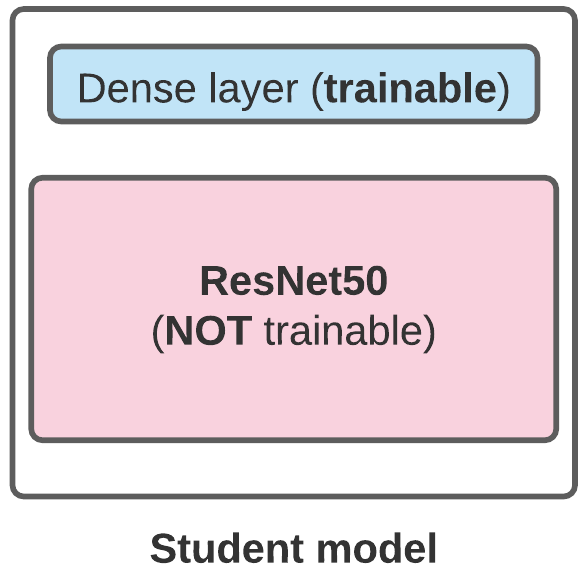}
\caption{Student model: the base network is frozen (not trainable); only the top dense layers are trained.}
\label{fig:student-model}
\end{figure}

For student model, we experimented with ResNet50 and VGG16 with dense layers added on top of these frozen models. In other words, we only train the dense layers as depicted in figure \ref{fig:student-model}

\subsubsection{\textbf Hypothesis}

In this experiment, we re-evaluate the effect of student architecture in knowledge distillation via teacher-forcing. Specifically, we want to experiment if student has better performance when it has the same architecture (but with much smaller size) with its teacher.

\begin{table}[h!]
\centering
\begin{tabular}{ |c|c|c| } 
 \hline
 \textbf{Student} & \textbf{SimCLR Teacher} & \textbf{Accuracy} \\
 \hline\hline
 VGG16 & ResNet50 & 0.65                                         \\
 ResNet50 & ResNet50 & 0.71                                      \\
 \hline
\end{tabular}
\caption{Teacher-student architecture effect on knowledge distillation.}
\label{table:teacher-student-architecture}
\end{table}

\subsubsection{\textbf Experiment Results and Analysis}

We trained student model with VGG16 and ResNet50 architecture. With teacher and student have the same architecture (ResNet50), the result is better than when teacher and student have different architecture as recorded in table \ref{table:teacher-student-architecture} 

Some hyper-parameters are included in table \ref{table:parameter}.

\begin{table}[h!]
\centering
\begin{tabular}{ |c|c| } 
 \hline
 \textbf{Parameter} & \textbf{Value} \\
 \hline\hline
 Epoch & 100                                    \\
 Learning rate & 0.01                                      \\
 Momentum & 0.9 \\
 Weight decay & 0.0001 \\
 Temperature &  1.0 \\
 \hline
\end{tabular}
\caption{Hyper-parameters}
\label{table:parameter}
\end{table}

The results support our conjecture that when student and teacher have the same architecture, the performance is better. It seems that it is easier for student to learn from teacher with the same architecture.

Interestingly, \cite{jang2019bharath} shows that learning from accurate teachers may be hard. Thus, we had to stop the training when it reaches the best accuracy as suggested by \cite{jang2019bharath}.

\subsection{Transfer Learning with SimCLR for Different Image Classification Tasks}

In this experiment, we evaluate the performance of SimCLR which was trained on ImageNet when it is applied to other datasets and whether the model can perform well in transfer learning settings. We also want to see whether we can reproduce the results mentioned in the SimCLR paper \cite{chen2020simple}.

\subsubsection{\textbf Experiment set-up}
This experiment is set up similar to experiment 6.1 although there are some key differences below: 

\begin{itemize}
  \item Dataset: we replaced Flower dataset with other TensorFlow datasets as listed in Table \ref{table:transfer-learning-accuracy}: Cats and dogs \cite{parkhi2012catsanddogs}, Oxford 102 Flowers \cite{nilsback2008flowers}, the Describable Textures Dataset (DTD) \cite{cimpoi2014dtd},  CIFAR-100 \cite{Krizhevsky2012cifar100} and Food-101 dataset \cite{bossard2014food101}.
  \item We applied random cropping with resize and flipping. We did not perform color augmentation or blurring. 
  \item We selected learning rate of 0.1, no weight decay and experiment with different batch sizes: 32, 64. We then fine-tuned the new linear layers for 30 iterations
  \item We ran the experiments on Google Colab infrastructure with TPU hardware accelerator and High-RAM. Even with this setting, the maximum batch size we were able to run was 64. Batch size larger than that caused the system to crash due to out of memory error. 
\end{itemize}

\subsubsection{\textbf Hypothesis}
\begin{itemize}
\item We conjecture that contrastive-based learning method would perform better when there is less number of classes, similar to how a 2-year old child can easily tell dog from cat, but would have trouble telling the difference among the 50 things that are presented to him or her. 
\item We also conjecture that transfer learning using SimCLR model also benefits from larger batch size since the training accuracy benefits from larger batch size.
\end{itemize}

\subsubsection{\textbf Experiment results and analysis}
Results in table \ref{table:transfer-learning-accuracy} show that indeed the number of classes is inversely correlated with final transfer accuracy after fine-tuning. One reason behind this could be explained by the fact that SimCLR learns representation by maximizing agreement between differently augmented views of the same data example via a contrastive loss in the latent space. With a large number of classes, the augmented images among different classes might not be easily distinguished any more. 

\begin{table}[h!]
\begin{center}
\begin{tabular}{ |c|c|c|c| } 
 \hline
 Dataset & Classes & Image Size & Accuracy   \\ 
 \hline\hline
 
 cats\_vs\_dogs & 2 & Varied & 1.0          \\
 tf\_flowers & 5 & Varied (~250x500) & 0.906    \\ 
 dtd & 47 & Varied (~300x640) & 0.656                     \\ 
 cifar\_100 & 100 & 32 x 32 & 0.25            \\ 
 food\_101 & 101 & Varied (512 x 512) & 0.125            \\ 
\hline
\end{tabular}
\caption{Transfer learning accuracy (batch size 64)}
\label{table:transfer-learning-accuracy}
\end{center}
\end{table}

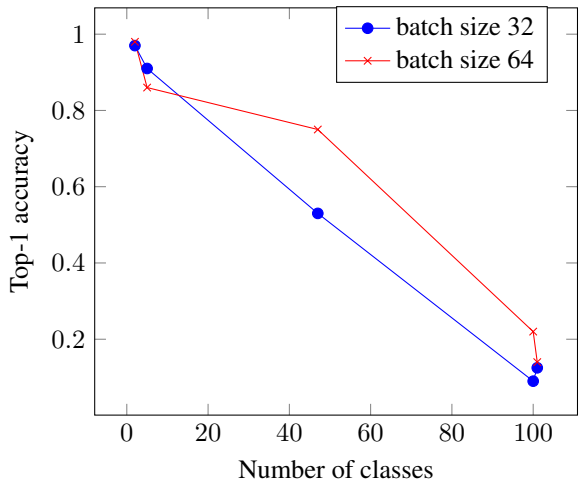
\begin{figure}[h]
\centering
\begin{tikzpicture}
\begin{axis}[
	xlabel=Number of classes,
	ylabel=Top-1 accuracy,
	width=8cm,height=7cm,
    legend style={at={(0.5,.91)},anchor=west}]

\addplot[color=blue,mark=*] coordinates {
	(2, 0.97)
	(5, 0.91)
	(47, 0.53)
	(100, 0.09)
	(101, 0.125)
};

\addplot[color=red,mark=x] coordinates {
	(2, 0.98)
	(5, 0.86)
	(47, 0.75)
	(100, 0.22)
	(101, 0.14)
};
\legend{batch size 32,batch size 64}
\end{axis}
\end{tikzpicture}
\caption{Effect of number of classes and batch size on Top-1 transfer accuracy}
\label{fig:transfer-learning-graph}
\end{figure}

With respect to batch size, figure \ref{fig:transfer-learning-graph} shows that datasets with more number of classes do benefit from larger batch size. However, due to the limitation of training in Colab environment, where we have limited RAM and TPUs, we could not train the data with batch size larger than 64 without running into memory error. Also due to this, we are not able to reproduce the higher accuracy quoted in SimCLR paper, which ran transfer learning with batch size = 256 for 200,000 steps for each dataset. 

\section{Discussion}
\subsection{Why bigger model is not overfitting when fine-tuning with fewer labels?}

One of the findings in the SimCLR paper is bigger models yield larger gains when fine-tuning with fewer label examples \cite{chen2020big}. At first, we thought that this finding is a little bit counter-intuitive, since bigger model would seem to be over-fitting on fewer label examples. But this might not be the case in this situation. We think that when the model gets bigger (more parameters), it will have many features to use for classification task. Then, when we fine-tune it on fewer label examples, it will have a greater search space for that one distinguished feature (or features) to be found.  For instance, when the bigger model is fine-tuned on 1\% of label examples, it will be forced to be exceptionally selective on the features it chose to classify, and thus results in larger gain when compared to fine-tuning on 100\% of label examples. 

\subsection{Loss Function for Knowledge Distillation}

In SimCLR, there are two different loss functions being used for each step of the process.

In self-supervised pretraining, the loss function is called NT-Xent loss or contrastive loss \cite{chen2020big}. This loss function takes into account the cosine similarity between two representation vectors of positive pairs (two augmented images from the same image), and two representation vectors of negative pairs (two augmented images from different images). It also uses temperature scalar to smoothen out the loss. The main purpose of this loss function is to ensure that similar augmented images are closer together in the latent space, and further away from dissimilar ones. 

In knowledge distillation of student-teacher architecture, the loss function is called distillation loss. This loss function is simple in a sense that instead of looking at the ground truth labels, the student model is learning from the teacher imputed labels. In instances where number of labels are significant, distillation loss can be altered slightly by adding a weighted factor to combine both the ground-truth labels and teacher labels for the student to learn from.


\subsection{Student Model in Experiment 6.2} 
As mentioned in experiment 6.2, we used ResNet50 and VGG16 architectures for the student model. However, the weights were frozen and not trainable. On top of this, we added trainable dense layers.



\subsection{Future Work} 

Contrast learning with data augmentation is proven to work well when the image has a single dominant object. Studying how well contrast learning can generalize when the image has multiple objects could potentially widen the applicability of contrast learning method. 

Designing a student architecture that is small enough but still can converge quickly and is able to match teacher's accuracy in the semi-supervised domain is still an area with little research. Future work could focus on this research area.

\newpage
\newpage

{\small
\bibliographystyle{ieee_fullname}
\bibliography{egbib}
}

\end{document}